\documentclass[runningheads]{llncs}

\usepackage{graphicx}
\usepackage{comment}
\usepackage{amsmath,amssymb} 
\usepackage{color}

\usepackage{epsfig}
\usepackage{multicol}
\usepackage{makecell}
\usepackage{subfigure}
\usepackage{multirow}

\usepackage[width=122mm,left=12mm,paperwidth=146mm,height=193mm,top=12mm,paperheight=217mm]{geometry}

\begin{document}
\pagestyle{headings}
\mainmatter

\title{Disentangled Non-Local Neural Networks}

\titlerunning{Disentangled Non-Local Neural Networks}
%
\author{Minghao Yin\inst{1}\thanks{Equal contribution. This work is done when Minghao Yin and Zhuliang Yao are interns at MSRA.} \and
Zhuliang Yao\inst{1,2\star} \and
Yue Cao\inst{2}  \and Xiu Li\inst{1} \and Zheng Zhang\inst{2} \and Stephen Lin\inst{2} \and Han Hu\inst{2}}
\authorrunning{Yin et al.}
%
\institute{Tsinghua University\\
\email{\{yinmh17,yzl17\}@mails.tsinghua.edu.cn}~~\email{ li.xiu@sz.tsinghua.edu.cn} \and
Microsoft Research Asia\\
\email{\{yuecao,zhez,stevelin,hanhu\}@microsoft.com}}
\maketitle


\begin{abstract}

The non-local block is a popular module for strengthening the context modeling ability of a regular convolutional neural network. This paper first studies the non-local block in depth, where we find that its attention computation can be split into two terms, a whitened pairwise term accounting for the relationship between two pixels and a unary term representing the saliency of every pixel. We also observe that the two terms trained alone tend to model different visual clues, e.g. the whitened pairwise term learns within-region relationships while the unary term learns salient boundaries. However, the two terms are tightly coupled in the non-local block, which hinders the learning of each. Based on these findings, we present the disentangled non-local block, where the two terms are decoupled to facilitate learning for both terms. We demonstrate the effectiveness of the decoupled design on various tasks, such as semantic segmentation on Cityscapes, ADE20K and PASCAL Context, object detection on COCO, and action recognition on Kinetics. Code is available at \\\url{https://github.com/yinmh17/DNL-Semantic-Segmentation}, \\ \url{https://github.com/Howal/DNL-Object-Detection}.

\end{abstract}

\section{Introduction}

The non-local block~\cite{wang2018non}, which models long-range dependency between pixels, has been widely used for numerous visual recognition tasks, such as object detection, semantic segmentation, and video action recognition. Towards better understanding the non-local block's efficacy, we observe that it can be viewed as a self-attention mechanism for pixel-to-pixel modeling. This self-attention is modeled as the dot-product between the features of two pixels in the embedding space. At first glance, this dot-product formulation represents \emph{pairwise} relationships. After further consideration, we find that it may encode \emph{unary} information as well, in the sense that a pixel may have its own independent impact on all other pixels. Based on this perspective, we split the dot-product based attention into two terms: a whitened pairwise term that accounts for the impact of one pixel {\em specifically} on another pixel, and a unary term that represents the influence of one pixel {\em generally} over all the pixels.

\begin{figure*}[t]
\begin{center}
\includegraphics[width=0.9\linewidth]{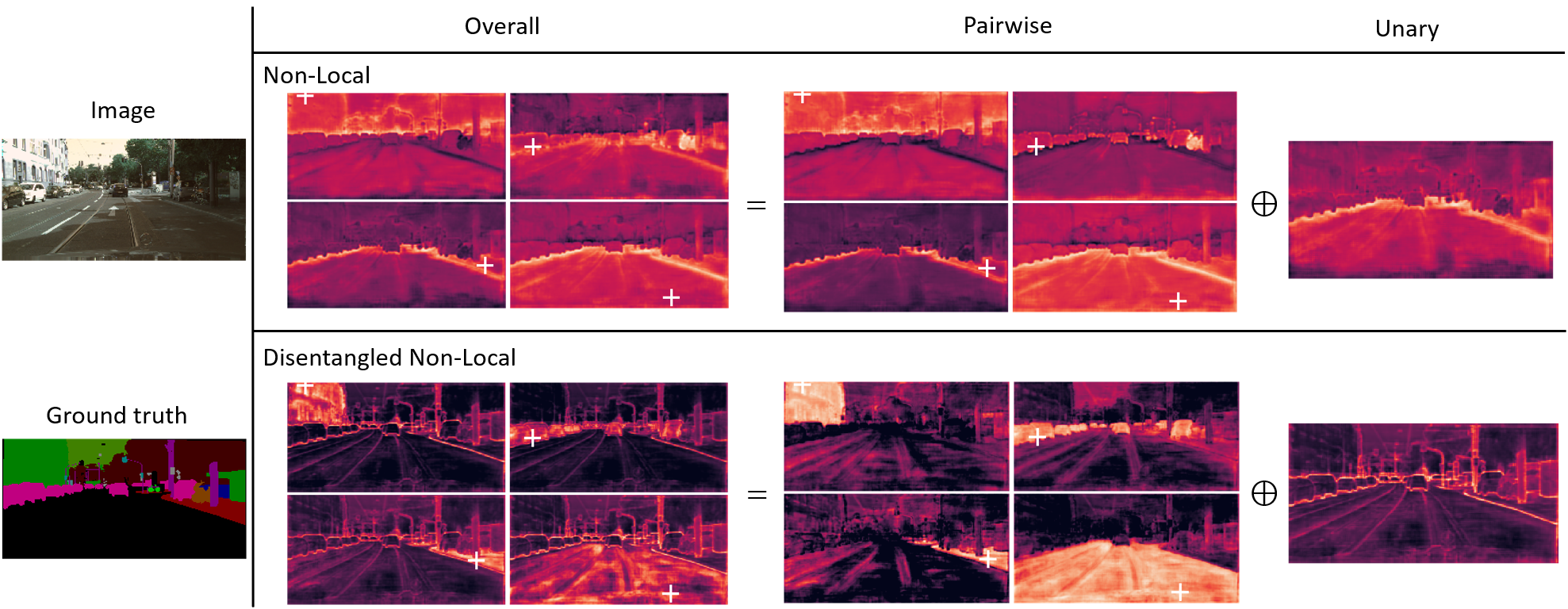}
\end{center}
\vspace{-20pt}
   \caption{Visualization of attention maps in the non-local block and our disentangled non-local block. With the disentanglement of our non-local block, the whitened pairwise term learns clear within-region clues while the unary term learns salient boundaries, which cannot be observed with the original non-local block}
\label{fig:teaser}
\vspace{-10pt}
\end{figure*}

We investigate the visual properties of each term without interference from the other. Specifically, we train two individual networks, with either the whitened pairwise term or the unary term removed in the standard attention formula of the non-local block. It is found that the non-local variant using the whitened pairwise term alone generally learns within-region relationships (the 2nd row of Fig.~\ref{fig:ablation}), while the variant using the unary term alone tends to model salient boundaries (the 3rd row of Fig.~\ref{fig:ablation}). However, the two terms do not learn such clear visual clues when they are both present within a non-local block, as illustrated in the top row of Fig.~\ref{fig:teaser}. This observation is verified via statistical analysis on the whole validation set. Also, the standard non-local block combining both terms performs even worse than the variant that includes only the unary term (shown in Table~\ref{tab:ablation}). This indicates that coupling the two terms together may be detrimental to the learning of these visual clues, and consequently affects the learning of discriminative features.

To address this problem, we present the disentangled non-local (DNL) block, where the whitened pairwise and unary terms are cleanly decoupled by using independent \textit{Softmax} functions and embedding matrices. With this disentangled design, the difficulty in joint learning of the whitened pairwise and unary terms is greatly diminished. As shown in second row of Fig.~\ref{fig:teaser}, the whitened pairwise term learns clear within-region clues while the unary term learns salient boundaries, even more clearly than what is learned when each term is trained alone.

The disentangled non-local block is validated through various vision tasks. On semantic segmentation benchmarks, by replacing the standard non-local block with the proposed DNL block with all other settings unchanged, significantly greater accuracy is achieved, with a 2.0\% mIoU gain on the Cityscapes validation set, 1.3\% mIoU gain on ADE20k, and 3.4\% on PASCAL-Context using a ResNet-101 backbone. With few bells and whistles, our DNL obtains state-of-the-art performance on the challenging ADE20K dataset. Also, with a task-specific DNL block, noticeable accuracy improvements are observed on both COCO object detection and Kinetics action recognition.

\section{Related Works}

\noindent \textbf{Non-local/self-attention.} These terms may appear in different application domains, but they refer to the same modeling mechanism. This mechanism was first proposed and widely used in natural language processing~\cite{britz2017massive,vaswani2017attention} and physical system modeling~\cite{watters2017visual,hoshen2017vain,santoro2017simple}. The self-attention / relation module affects an individual element (e.g. a word in a sentence) by aggregating features from a set of elements (e.g. all the words in the sentence), where the aggregation weights are usually determined on embedded feature similarities among the elements. They are powerful in capturing long-range dependencies and contextual information.

In the computer vision, two pioneering works~\cite{hu2017relation,wang2018non} first applied this kind of modeling mechanism to capture the relations between objects and pixels, respectively.
Since then, such modeling methods have demonstrated great effectiveness in many vision tasks, such as image classification~\cite{hu2019local}, object detection~\cite{hu2017relation,gu2018learning}, semantic segmentation~\cite{yuan2018ocnet}, video object detection~\cite{Deng_2019_ICCV,Wu_2019_ICCV,Guo_2019_ICCV,Chen_2020_CVPR} and tracking~\cite{Xu_2019_ICCV}, and action recognition~\cite{wang2018non}. There are also works that propose improvements to self-attention modeling, e.g. an additional relative position term~\cite{hu2017relation,hu2019local}, an additional channel attention~\cite{jun2019danet}, simplification~\cite{cao2019gcnet}, and speed-up~\cite{huang2019ccnet}.

This paper also presents an improvement over the basic self-attention / non-local neural networks. However, our work goes beyond straightforward application or technical modification of non-local networks in that it also brings a new perspective for understanding this module.

\noindent \textbf{Understanding non-local/self-attention mechanisms.} Our work is also related to several approaches that analyze the non-local/self-attention mechanism in depth, including the performance of individual terms~\cite{hu2017relation,tang2018analysis,zhu2019empirical} on various tasks. Also, there are studies which seek to uncover what is actually learnt by the non-local/self-attention mechanism in different tasks~\cite{cao2019gcnet}.

This work also targets a deeper understanding of the non-local mechanism, in a new perspective. Beyond improved understanding, our paper presents a more effective module, the disentangled non-local block, that is developed from this new understanding and is shown to be effective on multiple vision tasks.

\section{Non-local Networks in Depth}

\subsection{Dividing Non-local Block into Pairwise and Unary Terms}

Non-local block~\cite{wang2018non} computes pairwise relations between features of two positions to capture long-range dependencies. With $\mathbf{x}_i$ representing the input features at position $i$, the output features $\mathbf{y}_i$ of a non-local block are computed as
\begin{footnotesize}
\begin{equation}
\mathbf{y}_i=\sum_{j \in \Omega }\omega(\mathbf{x}_i,\mathbf{x}_j)g\left(\mathbf{x}_j\right),
\end{equation}
\end{footnotesize}where $\Omega$ denotes the set of all pixels on a feature map of size $H\times W$; $g(\cdot)$ is the \emph{value} transformation function with parameter $W_v$; $\omega(\mathbf{x}_i, \mathbf{x}_j)$ is the embedded similarity function from pixel $j$ (referred to as a \emph{key} pixel) to pixel $i$ (referred to as a \emph{query} pixel), typically instantiated by an Embedded Gaussian as
\begin{footnotesize}
\begin{equation}
\label{eq.nl}
\omega(\mathbf{x}_i,\mathbf{x}_j)=\sigma\left(\mathbf{q}_i^T\mathbf{k}_j\right)=\frac{\exp\left(\mathbf{q}_i^T\mathbf{k}_j\right)}{\sum_{t\in \Omega} \exp\left(\mathbf{q}_i^T\mathbf{k}_t\right)},
\end{equation}
\end{footnotesize}where $\mathbf{q}_i=W_q\mathbf{x}_i$ and $\mathbf{k}_j=W_k\mathbf{x}_j$ denote the \emph{query} and \emph{key} embedding of pixel $i$ and $j$, respectively, and $\sigma(\cdot)$ denotes the softmax function.

\begin{figure*}[t]
\centering
\includegraphics[width=0.9\linewidth]{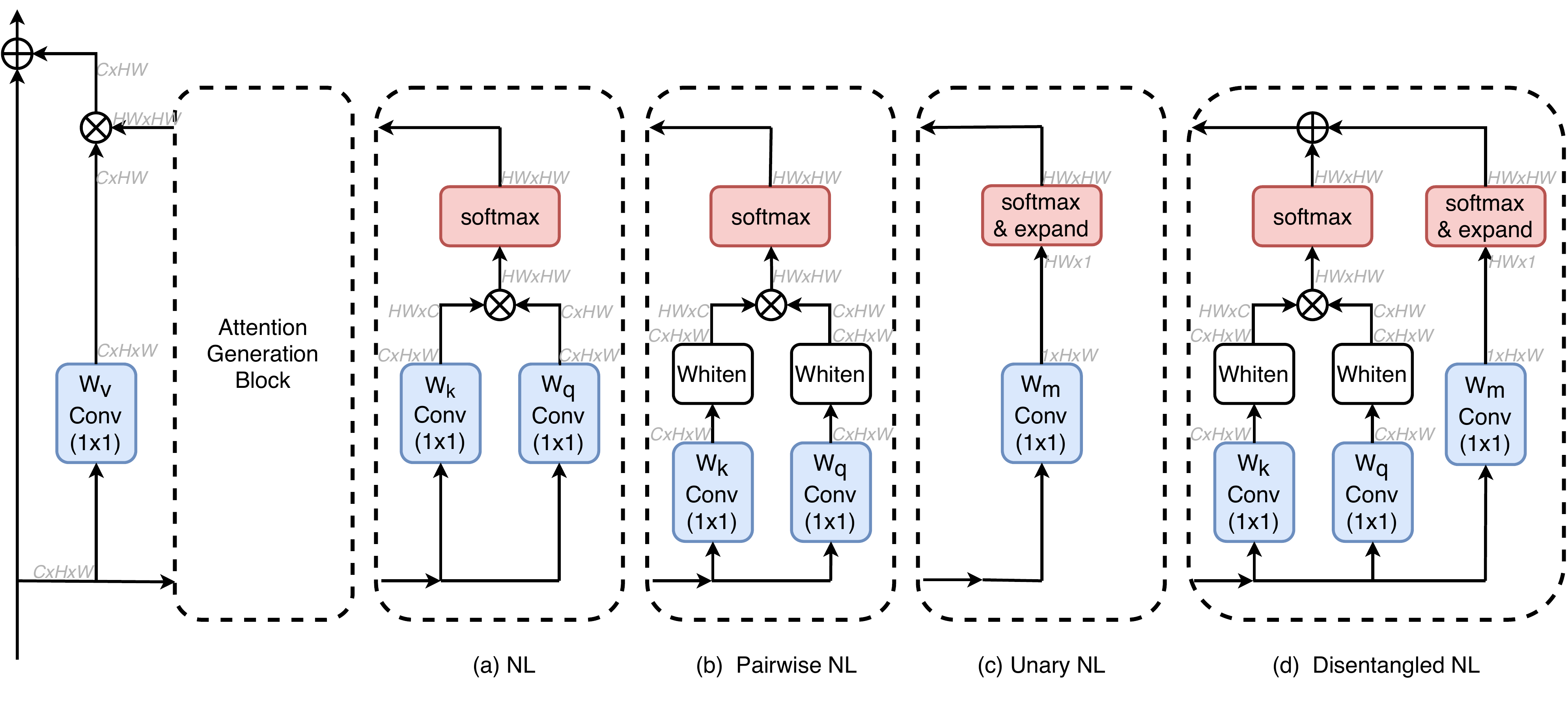}
\vspace{-15pt}
\caption{Architectures of non-local block, disentangled non-local block, and other variants. The shapes of feature maps are indicated in gray, \textit{e.g.}, {C$\times$H$\times$W}. ``$\otimes$" denotes matrix multiplication and ``$\oplus$" denotes element-wise addition. Blue boxes represent 1$\times$1 convolution. \textit{Softmax} is performed on the first dimension of feature maps}
\label{fig:dnl_block}
\vspace{-10pt}
\end{figure*}

At first glance, $\omega(\mathbf{x}_i, \mathbf{x}_j)$ (defined in Eq.~\ref{eq.nl}) appears to represent only a \emph{pairwise} relationship in the non-local block, through a dot product operation. However, we find that it may encode some \emph{unary} meaning as well. Considering a special case where the query vector is a constant over all image pixels, a \emph{key} pixel will have global impact on all \emph{query} pixels. In \cite{cao2019gcnet}, it was found that non-local blocks frequently degenerate into a pure \emph{unary} term in several image recognition tasks where each \emph{key} pixel in the image has the same similarity with all \emph{query} pixels. These findings indicate that the \emph{unary} term does exist in the non-local block formulation. It also raises a question of how to divide Eq.~(\ref{eq.nl}) into \emph{pairwise} and \emph{unary} terms, which account for the impact of one \emph{key} pixel specifically on another \emph{query} pixel and the influence of one \emph{key} pixel generally over all the \emph{query} pixels, respectively.

To answer this question, we first present a \emph{whitened} dot product between \emph{key} and \emph{query} to represent the \emph{pure} pairwise term: $\left(\mathbf{q}_i-\boldsymbol{\mu}_q\right)^T\left(\mathbf{k}_j-\boldsymbol{\mu}_k\right)$,
where $\boldsymbol{\mu}_q=\frac{1}{|\Omega|}\sum_{i \in \Omega}\mathbf{q}_i$ and $\boldsymbol{\mu}_k=\frac{1}{|\Omega|}\sum_{j \in \Omega}\mathbf{k}_j$ are the averaged \emph{query} and \emph{key} embedding over all pixels, respectively.
To remove the \emph{unary/global} component of \emph{key} pixels, the \emph{whitened} dot product is determined by maximizing the normalized differences between \emph{query} and \emph{key} pixels. In following proposition, we show how this can be achieved via an optimization objective, which allows for the whitened dot product to be computed.

\noindent \textbf{Proposition 1}: $\mathbf{\alpha}^* = \frac{1}{|\Omega|}\sum_{i \in \Omega}\mathbf{q}_i$, $\mathbf{\beta}^* = \frac{1}{|\Omega|}\sum_{m \in \Omega}\mathbf{k}_m$ is the optimal solution of the following optimization objective:
\begin{footnotesize}
\begin{equation}
\begin{aligned}
\arg \max_{\mathbf{\alpha},\mathbf{\beta}} & \;\;\; \frac{\sum_{i,m,n \in \Omega} \left((\mathbf{q}_i-\mathbf{\alpha})^T(\mathbf{k}_m-\mathbf{\beta})-(\mathbf{q}_i-\mathbf{\alpha})^T (\mathbf{k}_n -\mathbf{\beta})\right)^2}{\sum_{i \in \Omega} \left((\mathbf{q}_i-\mathbf{\alpha})^T (\mathbf{q}_i-\mathbf{\alpha})\right) \cdot \sum_{m,n \in \Omega} \left( (\mathbf{k}_m-\mathbf{k}_n)^T (\mathbf{k}_m-\mathbf{k}_n)\right) }   \\
& \;\;\;+ \frac{\sum_{m,i,j \in \Omega} \left((\mathbf{k}_m-\mathbf{\beta})^T(\mathbf{q}_i-\mathbf{\alpha})-(\mathbf{k}_m-\mathbf{\beta})^T (\mathbf{q}_j -\mathbf{\alpha})\right)^2}{\sum_{m \in \Omega} \left((\mathbf{k}_m-\mathbf{\beta})^T (\mathbf{k}_m-\mathbf{\beta})\right) \cdot \sum_{i,j \in \Omega} \left((\mathbf{q}_i-\mathbf{q}_j)^T (\mathbf{q}_i-\mathbf{q}_j)\right) }
\end{aligned}
\end{equation}
\end{footnotesize}

\noindent \textbf{Proof sketch}: The Hessian of the objective function $O$ with respect to $\alpha$ and $\beta$ is a non-positive definite matrix. The optimal $\alpha^*$ and $\beta^*$ are thus the solutions of the following equations: $\frac{\partial O} { \partial \alpha} = 0$, $\frac{\partial O} {\partial \beta} = 0$. Solving this yields $\mathbf{\alpha}^* = \frac{1}{|\Omega|}\sum_{i \in \Omega}\mathbf{q}_i$, $\mathbf{\beta}^* = \frac{1}{|\Omega|}\sum_{m \in \Omega}\mathbf{k}_m$. Please see the appendix for a detailed proof.

By extracting the whitened dot product as the \emph{pure} pairwise term, we can divide the dot product computation of the standard non-local block as
\begin{small}
\begin{align}
\mathbf{q}_i^T\mathbf{k}_j=\left(\mathbf{q}_i-\boldsymbol{\mu}_q\right)^T\left(\mathbf{k}_j-\boldsymbol{\mu}_k\right)+\boldsymbol{\mu}_q^T\mathbf{k}_j+\mathbf{q}_i^T\boldsymbol{\mu}_k+\boldsymbol{\mu}_q^T\boldsymbol{\mu}_k.
\end{align}
\end{small}

\begin{figure*}[t]
\begin{center}
\includegraphics[width=0.85\linewidth]{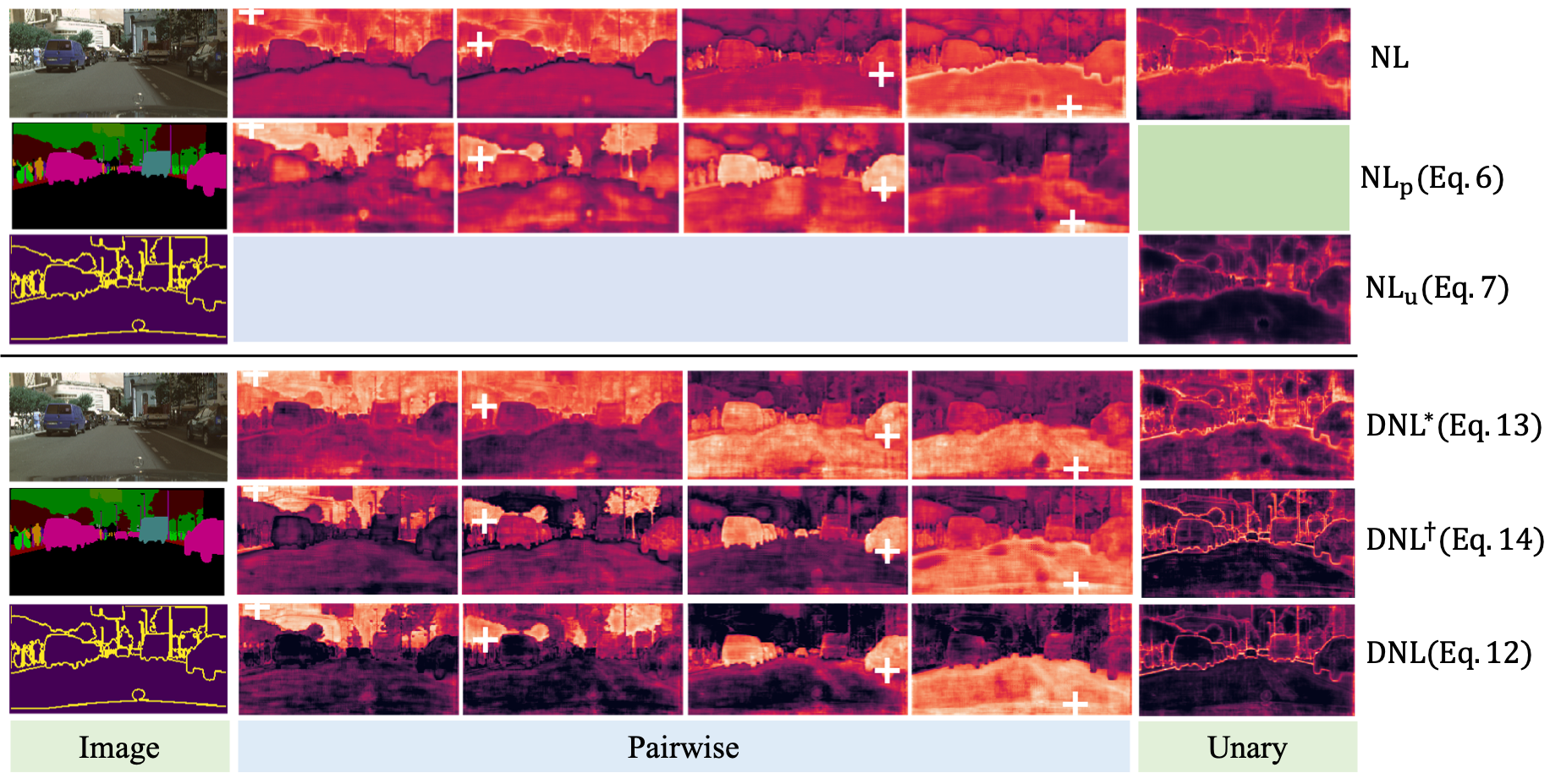}
\end{center}
\vspace{-20pt}
\caption{Visualization of attention maps for all variants of the NL block mentioned in this paper. Column 1: image, ground truth and edges of ground truth. Columns 2-5: attention maps of pairwise terms. Column 6: attention maps of unary terms. As NL$_{u}$ has no pairwise attention map, and NL$_{p}$ has no unary attention map, we leave the corresponding spaces empty}
\label{fig:ablation}
\vspace{-10pt}
\end{figure*}

Note that the last two terms ($\mathbf{q}_i^T\boldsymbol{\mu}_k$ and $\boldsymbol{\mu}_q^T\boldsymbol{\mu}_k$) are factors that appear in both the numerator and denominator of Eq.~(\ref{eq.nl}). Hence, these two terms can be eliminated (see proof in the Appendix). After this elimination, we reach the following \emph{pairwise} and \emph{unary} split of a standard non-local block:
\begin{small}
\begin{align}
\label{eq.nl_split}
\omega(\mathbf{x}_i,\mathbf{x}_j)=\sigma(\mathbf{q}_i^T\mathbf{k}_j)=\sigma(\underbrace{\left(\mathbf{q}_i-\boldsymbol{\mu}_q\right)^T\left(\mathbf{k}_j-\boldsymbol{\mu}_k\right)}_{\text{pairwise}}+\underbrace{\boldsymbol{\mu}_q^T\mathbf{k}_j}_{\text{unary}}),
\end{align}
\end{small}
where the first \emph{whitened} dot product term represents the \emph{pure} pairwise relation between a \emph{query} pixel $i$ and a \emph{key} pixel $j$, and the second term represents the \emph{unary} relation where a \emph{key} pixel $j$ has the same impact on all \emph{query} pixels $i$.

\subsection{What Visual Clues are Expected to be Learnt by Pairwise and Unary Terms?}

\label{sec.visual_clues}

\begin{table}[t]
\small
\centering
\caption{Consistency statistics between attention maps of the non-local variants and the ground-truth within-category and boundary maps on the Cityscapes validation set}
\scalebox{0.9}{
\begin{tabular}{l|c|c|c}
\Xhline{1.0pt}
method & pair $\cap$ within-category & pair $\cap$ boundary &  unary $\cap$ boundary \\
\hline
random & 0.259 & 0.132 & 0.135\\
\hline
pairwise NL (Eq.~\ref{con:nlp}) & 0.635 & 0.141  & - \\
unary NL (Eq.~\ref{con:nlu}) & - & - & 0.460 \\
\hline
NL (Eq.~\ref{eq.nl}) &0.318 & 0.160 & 0.172 \\
\hline
DNL$^*$ (Eq.~\ref{eq.nl_variant1}) & 0.446 & 0.146 & 0.305\\
DNL$^\dag$ (Eq.~\ref{eq.nl_variant2}) & 0.679 & 0.137 & 0.657 \\
\hline
DNL (Eq.~\ref{eq.dnl}) & 0.759 & 0.130 & 0.696 \\
\Xhline{1.0pt}
\end{tabular}
}
\label{tab:stat_city}
\vspace{-20pt}
\end{table}

To study what visual clues are expected to be learnt by the pairwise and unary terms, respectively, we construct two variants of the non-local block by using either the pairwise or unary term alone, such that the influence of the other term is eliminated. The two variants use the following similarity computation functions instead of the one in Eq.~(\ref{eq.nl}):
\begin{small}
\begin{align}
\omega_{\text{p}}\left({\mathbf x}_i,{\mathbf x}_j\right)&=\sigma\left(\left(\mathbf{q}_i-\boldsymbol{\mu}_q\right)^T\left(\mathbf{k}_j-\boldsymbol{\mu}_k\right)\right), \label{con:nlp}\\
\omega_{\text{u}}\left({\mathbf x}_i,{\mathbf x}_j\right)&=\sigma(\boldsymbol\mu_q^T{\mathbf k}_j). \label{con:nlu}
\end{align}
\end{small}
The two variants are denoted as ``pairwise NL'' and ``unary NL'', and illustrated in Fig.~\ref{fig:dnl_block}(b)  and \ref{fig:dnl_block}(c), respectively. We apply these two variants of non-local block to the Cityscapes semantic segmentation~\cite{cordts2016cityscapes} (see Section~\ref{sec.semantic_seg} for detailed settings), and visualize their learnt attention (similarity) maps on several randomly selected validation images in Cityscapes, as shown in Fig.~\ref{fig:ablation} (please see more examples in the Appendix). It can be seen that the pairwise NL block tends to learn pixel relationships within the same category region, while the unary NL block tends to learn the impact from boundary pixels to all image pixels.

This observation is further verified by quantitative analysis using the ground-truth region and boundary annotations in Cityscapes. Denote $P^{(i)}$=$\{\omega _\text{p}(\mathbf{x}_i, \mathbf{x}_j) |$ $j \in \Omega\} \in \mathbb{R}^{H\times W}$ as the attention map of pixel $i$ according to the pairwise term of Eq.~(\ref{con:nlp}), $U$=$\{\omega _\text{u}(\mathbf{x}_i, \mathbf{x}_j) | j \in \Omega\} \in \mathbb{R}^{H\times W}$ as the attention map for all query pixels according to the unary term of Eq.~(\ref{con:nlu}), $C^{(i)} \in \mathbb{R}^{H\times W}$ as the binary within-category region map of pixel $i$, and $E \in \mathbb{R}^{H\times W}$ as the binary boundary map indicating pixels with distance to ground truth contour of less than 5 pixels.

We evaluate the consistency between attention maps $A \in \{P^{(i)}, U\}$ and ground-truth boundary/same-category region $G \in \{C^{(i)}, E\}$ by their overlaps:
\begin{equation}
    A \cap G =\sum_{j \in \Omega} A_j \odot G_j,
\end{equation}
where $A_j, G_j$ are the element values of the corresponding attention map and binary map at pixel $j$, respectively.

Table~\ref{tab:stat_city} shows the averaged consistency measures of the attention maps in Eq.~(\ref{con:nlp}) and Eq.~(\ref{con:nlu}) to ground-truth region maps (denoted as pairwise NL and unary NL) using all 500 validation images in the Cityscapes datasets. We also report the consistency measures by a random attention map for reference (denoted as random). The following can be seen:
\begin{itemize}
    \item The attention map by the pairwise NL block of Eq.~(\ref{con:nlp}) has significantly larger overlap with the ground-truth same-category region than the random attention map (0.635 vs. 0.259), but has similar overlap with the ground-truth boundary region (0.141 vs. 0.132), indicating that \emph{the pure pairwise term tends to learn relationship between pixels within same-category regions}.
    \item The attention map by the unary NL block of Eq.~(\ref{con:nlu}) has significantly larger overlap with the ground-truth boundary region than the random attention map (0.460 vs. 0.135), indicating that \emph{the unary term tends to learn the impact of boundary pixels on all image pixels}. This is likely because the image boundary area provides the most informative cues when considering the general effect on all pixels.
\end{itemize}

\subsection{Does the Non-local Block Learn Visual Clues Well?}

We then study the learnt pairwise and unary terms by the non-local block. We follow Eq.~(\ref{eq.nl_split}) to split the standard similarity computation into the pairwise and unary terms, and normalize them by a softmax operation separately. After splitting and normalization, we can compute their overlaps with the ground-truth within-category region map and boundary region map, as shown in Table~\ref{tab:stat_city}.

It can be seen that the pairwise term in the standard NL block which is jointly learnt with the unary term has significantly smaller overlap with the ground-truth within-category region than in the pairwise NL block where the pairwise term is learnt alone (0.318 vs. 0.635). It can be also seen that the unary term in the standard NL block which is jointly learnt with the pairwise term has significantly smaller overlap with the ground-truth boundary region than in the unary NL block where the unary term is learnt alone (0.172 vs. 0.460). These results indicate that neither of the pairwise and unary terms learn the visual clues of within-category regions and boundaries well, as also demonstrated in Fig.~\ref{fig:teaser} (top).

\subsection{Why the Non-Local Block Does Not Learn Visual Clues Well?}

To understand why the non-local block does not learn the two visual clues well, while the two terms alone can clearly learn them, we rewrite Eq.~(\ref{eq.nl_split}) as:
\begin{small}
\begin{align}
\label{eq.rewrite_nl}
\sigma(\mathbf{q}_i\cdot\mathbf{k}_j)&=\sigma\left(\left(\mathbf{q}_i-\mathbf{\boldsymbol{\mu}}_q\right)^T\left(\mathbf{k}_j-\boldsymbol{\mu}_k\right)+\boldsymbol{\mu}_q^T\mathbf{k}_j\right) \notag \\
&=\frac1{\lambda_i}\sigma\left(\left(\mathbf{q}_i-\boldsymbol{\mu}_q\right)^T\left(\mathbf{k}_j-\boldsymbol{\mu}_k\right)\right)\cdot\sigma(\boldsymbol{\mu}_q^T\mathbf{k}_j) = \frac1{\lambda_i} \omega_\text{p}(\mathbf{x}_i, \mathbf{x}_j) \cdot \omega_\text{u}(\mathbf{x}_i, \mathbf{x}_j),
\end{align}
\end{small}where $\lambda_i$ is a normalization scalar such that the sum of attention map values over $\Omega$ is 1.

Consider the back-propagation of loss $L$ to the pairwise and unary terms:
\begin{small}
\[
\frac{\partial L}{\partial \sigma(\omega_\text{p})}=\frac{\partial L}{\partial \sigma(\omega)} \cdot \frac{\partial \sigma(\omega)}{\partial \sigma(\omega_\text{p})}=\frac{\partial L}{\partial \sigma(\omega)} \cdot \sigma(\omega_{\text{u}}),
\]
\[
\frac{\partial L}{\partial \sigma(\omega_{\text{u}})}=\frac{\partial L}{\partial \sigma(\omega)} \cdot \frac{\partial \sigma(\omega)}{\partial \sigma(\omega_{\text{u}})}=\frac{\partial L}{\partial \sigma(\omega)} \cdot \sigma(\omega_{\text{p}}).
\]
\end{small}
It can be seen that both gradients are determined by the value of the other term. When the value of the other term becomes very small (close to 0), the gradient of this term will be also very small, thus inhibiting the learning of this term. For example, if we learn the unary term to well represent the boundary area, the unary attention weights on the non-boundary area will be close to 0 and the pairwise term at the non-boundary area would thus be hard to learn well due to the vanishing gradient issue. On the other hand, if we learn the pairwise term to well represent the within-category area, the unary attention weights on the boundary area will be close to 0 and the pairwise term at the non-boundary area would also be hard to learn well due to the same vanishing gradient issue.

Another problem is the \emph{shared} key transformation $W_k$ used in both the pairwise and unary terms, causing the computation of the two terms to be coupled. Such coupling may introduce additional difficulties in learning the two terms.

\section{Disentangled Non-local Neural Networks}

In this section, we present a new non-local block, named disentangled non-local (DNL) block, which effectively disentangles the learning of pairwise and unary terms. In the following sections, we first describe how we modify the standard non-local (NL) block into a disentangled non-local (NL) block, such that the two visual clues described above can be learnt well. Then we analyze its actual behavior in learning visual clues using the method in Section~\ref{sec.visual_clues}.

\subsection{Formulation}

\label{sec.dnl_formulation}

Our first modification is to change the \emph{multiplication} in Eq.~(\ref{eq.rewrite_nl}) to \emph{addition}:
\begin{small}
\begin{align}
&\omega(\mathbf{x}_i, \mathbf{x}_j) =  \omega_\text{p}(\mathbf{x}_i, \mathbf{x}_j) \cdot \omega_\text{u}(\mathbf{x}_i, \mathbf{x}_j) \;\;\Rightarrow \;\; \omega(\mathbf{x}_i, \mathbf{x}_j) = \omega_\text{p}(\mathbf{x}_i, \mathbf{x}_j) + \omega_\text{u}(\mathbf{x}_i, \mathbf{x}_j).
\end{align}
\vspace{-10pt}
\end{small}

The gradients of these two terms are
\begin{small}
\[
\frac{\partial L}{\partial \sigma(\omega_{\text{p}})}=\frac{\partial L}{\partial \sigma(\omega)}, \frac{\partial L}{\partial \sigma(\omega_{\text{u}})}=\frac{\partial L}{\partial \sigma(\omega)}.
\]
\end{small}
So the gradients of each term will not be impacted by the other.

The second modification is to change the transformation $W_k$ in unary term to be an independent linear transformation $W_m$ with output dimension of 1:
\begin{small}
\begin{align}
 \boldsymbol\mu_q^T{\mathbf k}_j=\boldsymbol\mu_q^TW_k\mathbf{x}_j\Rightarrow  m_j=W_m\mathbf{x}_j.
\end{align}
\end{small}After this modification, the pairwise and unary terms will no longer share the $W_k$ transformation, which further decouples them.

\noindent \textbf{DNL formulation.} With these two modifications, we obtain the following similarity computation for the disentangled non-local (DNL) block:
\begin{small}
\begin{align}
\label{eq.dnl}
\omega^{\text{D}} (\mathbf{x}_i, \mathbf{x}_j) = \sigma\left(\left(\mathbf{q}_i-\boldsymbol{\mu}_q\right)^T\left(\mathbf{k}_j-\boldsymbol{\mu}_k\right)\right) + \sigma(m_j).
\end{align}
\end{small}The resulting DNL block is illustrated in Fig.~\ref{fig:dnl_block} (d). Note that we adopt a single \emph{value} transform for both pairwise and unary terms, which is similarly effective on benchmarks as using independent value transform but with reduced complexity.

\noindent \emph{Complexity.} For an input feature map of $C\times H \times W$, we follow~\cite{wang2018non} by using $C/2$ dimensional \emph{key} and \emph{query} vectors. The space and time complexities are $\mathcal{O}^D(\text{space}) = (2C+1)C$ and $\mathcal{O}^D(\text{time}) = \left((2C+1)C + (\frac{3}{2}C+2)HW\right)HW$, respectively. For reference, the space and time complexity of a standard non-local block are $\mathcal{O}(\text{space}) = 2C^2$ and $\mathcal{O}(\text{time}) = \left(2C^2 + (\frac{3}{2}C+1)HW\right)HW$, respectively. The additional space and computational overhead of the disentangled non-local block over a standard non-local block is marginal, specifically 0.1\% and 0.15\% for $C=512$ in our semantic segmentation experiments.

\noindent \textbf{DNL variants for diagnostic purposes.} To diagnose the effects of the two decoupling modifications alone, we consider the following two variants:
\begin{small}
\begin{align}
& \omega^{\text{D}*} (\mathbf{x}_i, \mathbf{x}_j) = \sigma\left(\left(\mathbf{q}_i-\boldsymbol{\mu}_q\right)^T\left(\mathbf{k}_j-\boldsymbol{\mu}_k\right) + m_j\right), \label{eq.nl_variant1} \\
&\omega^{\text{D}\dag} (\mathbf{x}_i, \mathbf{x}_j) = \sigma\left(\left(\mathbf{q}_i-\boldsymbol{\mu}_q\right)^T\left(\mathbf{k}_j-\boldsymbol{\mu}_k\right)\right) + \sigma(\boldsymbol\mu_q^T \mathbf{k}_j), \label{eq.nl_variant2}
\end{align}
\end{small}which each involves only one of the two modifications.

\begin{figure*}[t]
    \centering
    \includegraphics[width=0.8\linewidth]{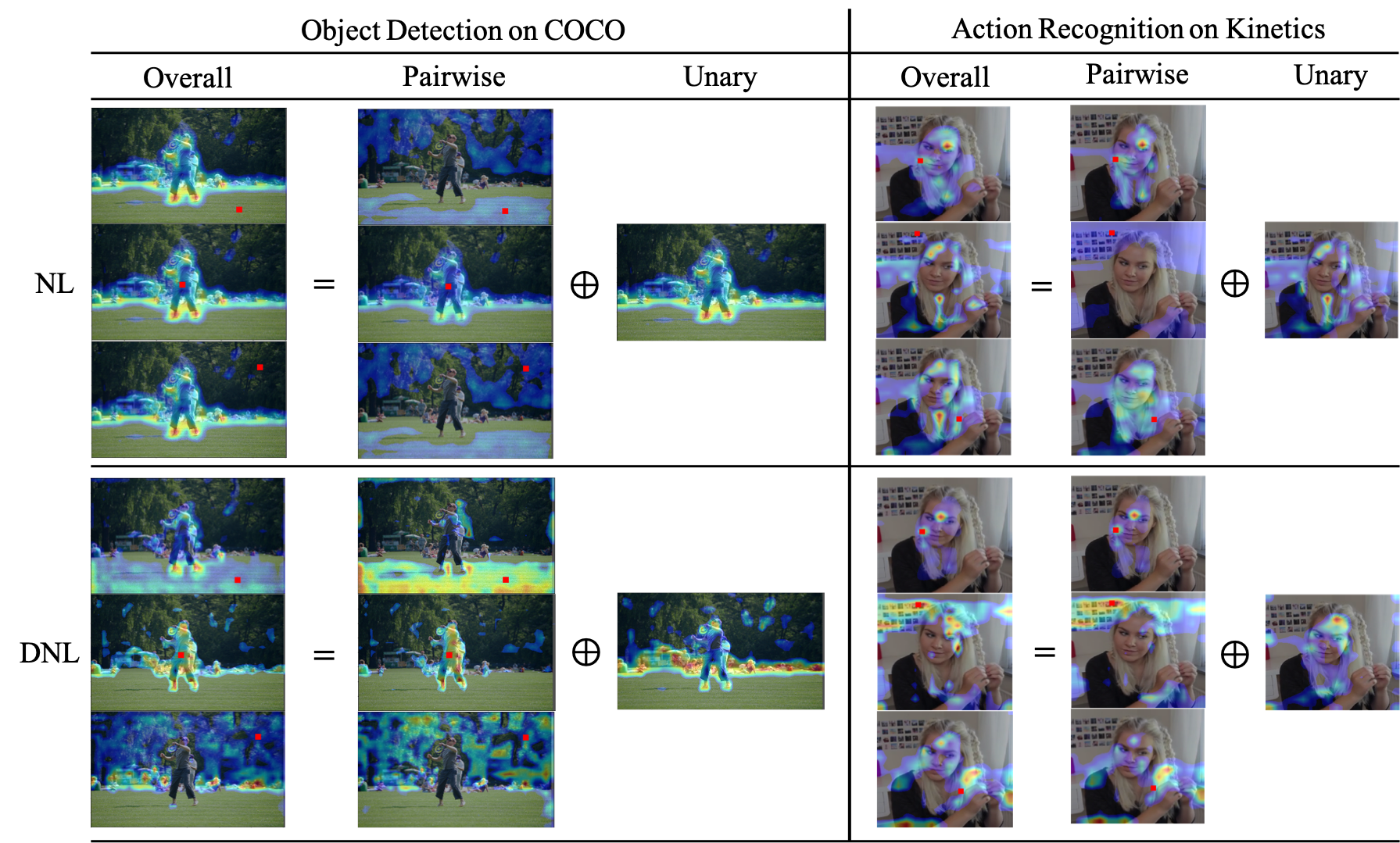}
    \vspace{-10pt}
    \caption{Visualization of attention maps in NL and our DNL block on COCO object detection and Kinetics action recognition. The query points are marked in red. Please refer to appendix for more examples}
    \label{fig:vis-coco-kinetics}
    \vspace{-15pt}
\end{figure*}

\subsection{Behavior of DNL on Learning Visual Clues }
\label{sec:behavior}
 We compute the overlaps of the pairwise and unary attention maps in DNL (Eq.~\ref{eq.dnl}) with the ground-truth within-category region map and boundary region map, as shown in Table~\ref{tab:stat_city}.

It can be seen that the pairwise term in DNL has significantly larger overlap with the ground-truth within-category region than the one in the standard NL block (0.759 vs. 0.318), and the unary term has significantly larger overlap with the boundary region than that in the standard NL block (0.696 vs. 0.172). These results indicate better learning of the two visual clues by the DNL block in comparison to the standard NL block.

Compared with the blocks which learn the pairwise or unary terms alone (see the ``pairwise NL'' and ``unary NL'' rows), such measures are surprisingly 0.124 and 0.236 higher with DNL. We hypothesize that when one term is learned alone, it may encode some portion of the other clue, as it is also useful for inference. By explicitly learning both terms, our disentangling design can separate one from the other, allowing it to better extract these visual clues.

We then verify the effects of each disentangling modification by these measures. By incorporating the ``disentangled transformation'' modification alone ($\omega^{*}$) as in Eq.~(\ref{eq.nl_variant1}), it achieves 0.446 and 0.305 on within-category modeling and boundary modeling, respectively, which is marginally better than the standard non-local block. By incorporating the ```multiplication to addition'' modification alone ($\omega^{\dag}$) as in Eq.~(\ref{eq.nl_variant2}), it achieves 0.679 and 0.657 on within-category modeling and boundary modeling, respectively.

The results indicate that the two modifications both benefit the learning of two visual clues and work better if combined together. The improvements in visual clue modeling by two disentangling strategies are also illustrated in Fig.~\ref{fig:ablation}.

Note such disentangling strategies also effect on other tasks beyond semantic segmentation. In object detection and action recognition tasks, we also observe clearer learnt visual clues by the DNL block than by the standard NL. As shown in Fig.~\ref{fig:vis-coco-kinetics}, while in NL the pairwise term is almost hindered by the unary term (also observed by~\cite{cao2019gcnet}), the pariwise term in DNL shows clear within-region meaning and appears significant in the final overall attention maps. The unary term in DNL also shows more focus to salient regions (not limited to boundaries which is different from that observed in the semantic segmentation task) than the one in an NL block. More examples will be shown in appendix.

\section{Experiments}

We evaluate the proposed DNL method on the recognition tasks of semantic segmentation, object detection/instance segmentation, and action recognition.

\subsection{Semantic Segmentation}

\label{sec.semantic_seg}

\textbf{Datasets.} We use three benchmarks for semantic segmentation evaluation.

\emph{Cityscapes}~\cite{cordts2016cityscapes} focuses on semantic understanding of urban street scenes. It provides a total of 5,000 finely annotated images, which is divided into 2,975/500/ 1,525 images for training, validation and testing. Additional 20,000 coarsely annotated images are also provided. The dataset contains annotations for over 30 classes, of which 19 classes are used in evaluation.

\emph{ADE20K}~\cite{zhou2017scene} was used in the ImageNet Scene Parsing Challenge 2016 and covers a wide range of scenes and object categories. It contains 20K images for training, 2K images for validation, and another 3K images for testing. It includes 150 semantic categories for evaluation.

\emph{PASCAL-Context}~\cite{mottaghi2014the} is a set of additional annotations for PASCAL VOC 2010, which label more than 400 categories of 4,998 images for training and 5,105 images for validation. For semantic segmentation, 59 semantic classes and 1 background class are used in training and validation.

\noindent \textbf{Architecture.} We follow recent practice~\cite{huang2019ccnet} by using dilated FCN~\cite{long2015fcn} and a ResNet101~\cite{he2016deep} backbone for our major segmentation experiments. The strides and dilations of $3\times 3$ convolutions are set to 1 and 2 for stage4, and 1 and 4 for stage5. The baseline model uses a segmentation head consisting of a $3\times3$ convolution layer to reduce the channels to 512 and a subsequent classifier to predict the final segmentation results. For experiments with a non-local or a disentangled non-local block, the block is inserted right before the final classifier.

\begin{table}[t]
\small
\centering
\caption{Ablation study on the validation set of Cityscapes}
\subfigure[Decoupling strategy]{
\scalebox{0.9}{
\begin{tabular}{l|c|c|c}
\Xhline{1.0pt}
 & mul $\rightarrow$ add & non-shared $W_k$  & mIoU \\
\hline
Baseline & - & - & 75.8 \\
\hline
NL & $\times$ & $\times$ & 78.5 \\
DNL$^\dag$(\ref{eq.nl_variant2}) & $\surd$ & $\times$ & 79.2 \\
DNL*(\ref{eq.nl_variant1})  & $\times$ & $\surd$ & 79.0 \\
DNL & $\surd$ & $\surd$ & 80.5 \\
\Xhline{1.0pt}
\end{tabular}
}
}
\subfigure[Pairwise and unary terms]{
\scalebox{0.9}{
\begin{tabular}{l|c|c|c}
\Xhline{1.0pt}
 & pairwise term & unary term & mIoU \\
\hline
Baseline & - & - & 75.8 \\
\hline
NL & $\surd$ & $\surd$ & 78.5 \\
NL$_{p}$ & $\surd$ & $\times$ & 77.5 \\
NL$_{u}$ & $\times$ & $\surd$ & 79.3 \\
DNL & $\surd$ & $\surd$ & 80.5 \\
\Xhline{1.0pt}
\end{tabular}
}
}
\vspace{0pt}
\label{tab:ablation}
\vspace{-20pt}
\end{table}

\noindent \textbf{Training and Inference.}
The implementation and hyper-parameters mostly follow~\cite{huang2019ccnet}. The SGD optimizer with poly learning rate policy ($1-(\frac{iter}{iter_{max}})^{0.9}$) is employed.
For Cityscapes, the networks are trained on 4 GPUs with 2 images per GPU for 60K iterations. The initial learning rate is 0.01, the weight decay is 0.0005. Input images are cropped to $769\times769$.
For ADE20K, the networks are trained on 8 GPUs with 2 images per GPU for 150K iterations. The initial learning rate is 0.02, and the weight decay is 0.0001. Input images are cropped to $520\times520$.
For PASCAL-Context, the network is trained on 4 GPUs with 4 images per GPU for 30K iterations. The initial learning rate is 0.001, and the weight decay is 0.0001. Input images are cropped to $520\times520$.
For all datasets, the data is augmented with random horizontal flipping, random scaling within $[0.5, 2.0]$, and random brightness jittering of $[-10, 10]$.
Following~\cite{yuan2018ocnet}, online hard example mining (OHEM) and an auxiliary loss on the output of conv4 with a weight of 0.5 are employed for Cityscapes and ADE20K, only auxiliary loss is employed for PASCAL-Context.

We mostly follow~\cite{huang2019ccnet} in inference. For Cityscapes, we sample $769\times 769$ windows for inference and their results are fused to generate the prediction of an entire image. For other datasets, we resize the image resolution to be the same as in training and a multi-scale test is adopted.

\begin{table}[t]
\small
\centering
    \addtolength{\tabcolsep}{6.5pt}
\caption{Comparisons with state-of-the-art approaches on the Cityscapes test set}
\scalebox{0.9}{
\begin{tabular}{l|c|c|c|c}
\Xhline{1.0pt}
Method & Backbone & ASPP & Coarse & mIoU (\%) \\
\hline
PSANet~\cite{zhao2018psanet} & ResNet-101 &  &  & 80.1 \\
DANet~\cite{jun2019danet} & ResNet-101 &  &  & 81.5 \\
HRNet~\cite{sun2019hrnet} & HRNetV2-W48 &  &  & 81.9 \\
SeENet~\cite{pang2019seenet} & ResNet-101 &  &  & 81.2 \\
SPGNet~\cite{cheng2019spgnet} & ResNet-101 &  &  & 81.1 \\
CCNet~\cite{huang2019ccnet} & ResNet-101 &  &  & 81.4 \\
ANN~\cite{zhu2019ann} & ResNet-101 &  &  & 81.3 \\
DenseASPP~\cite{yang2018denseaspp} & DenseNet-161 & $\surd$ &  & 80.6 \\
OCNet~\cite{yuan2018ocnet} & ResNet-101 & $\surd$ &  & 81.7 \\
ACFNet~\cite{zhang2019acfnet} & ResNet-101 & $\surd$ &  & 81.8 \\
PSPNet~\cite{zhao2017pspnet} & ResNet-101 &  & $\surd$ & 81.2 \\
PSANet~\cite{zhao2018psanet} & ResNet-101 &  & $\surd$ & 81.4 \\
DeepLabv3~\cite{chen2017deeplabv3} & ResNet-101 & $\surd$ & $\surd$ & 81.3 \\
\hline
NL & ResNet-101 &  & $\surd$ & 80.8 \\
DNL (ours) & ResNet-101 & & $\surd$ & 82.0 \\
\hline
NL & HRNetV2-W48 &  & $\surd$ & 82.5 \\
DNL (ours) & HRNetV2-W48 &  & $\surd$ & 83.0 \\
\Xhline{1.0pt}
\end{tabular}
}
\label{tab:sota_ccityscapes}
\vspace{-20pt}
\end{table}

\noindent \textbf{Ablation Study}
We ablate several design components in the proposed disentangled non-local block on the Cityscapes validation set. A ResNet-101 backbone is adopted for all ablations.

\noindent \emph{DNL variants.}
The disentangled non-local block has two decoupling modifications on the standard non-local block: multiplication to addition, and separate \emph{key} transformations. In addition to comparing the full DNL model with the standard non-local model, we also conduct experiments for these two variants which include only one of the decoupling modifications.

The results are shown in Table~\ref{tab:ablation}(a). While the standard
non-local model brings 2.7\% mIoU gains over a plain ResNet-101 model (78.5\% vs. 75.8\%), by replacing the standard non-local block by our disentangled non-local block, we achieve an additional 2.0\% mIoU gain over the standard non-local block (80.5\% vs. 78.5\%), with almost no complexity increase.

The variants that use each decoupling strategy alone achieve 0.5\% and 0.7\% mIoU gains over the standard non-local block (79.0 vs. 78.5 and 79.2 vs. 78.5), showing that both strategies are beneficial alone. They are also both crucial, as combining them leads to significantly better performance than using each alone.

\noindent \emph{Effects of pairwise and unary term alone.}
Table~\ref{tab:ablation}(b) compares the methods using the pairwise term or unary term alone. Using the pairwise term alone achieves 77.5\% mIoU, which is 1.7\% better than the baseline plain network without it. Using the unary term alone achieves 79.3\% mIoU, which is 3.5\% better than the baseline plain network and even 0.8\% mIoU better than the standard non-local network which models both pairwise and unary terms. These results indicate that the standard non-local block hinders the effect of the unary term, probably due to the coupling of two kinds of relationships. Our disentangled non-local networks effectively disentangle the two terms, and thus can better exploit their effects to achieve a higher accuracy of 80.5\% mIoU.

\begin{table}[t]
\small
\centering
    \addtolength{\tabcolsep}{2.5pt}
\caption{Comparisons with state-of-the-art approaches on the validation set and test set of ADE20K, and test set of PASCAL-Context}
\scalebox{0.9}{
\begin{tabular}{l|c|cc|c}
\Xhline{1.0pt}
\multirow{2}{*}{Method} & \multirow{2}{*}{Backbone}  & \multicolumn{2}{c|}{ADE20K} & PASCAL-Context \\
& & val mIoU (\%)  & test mIoU (\%) &  test mIoU (\%) \\
\hline
PSANet~\cite{zhao2018psanet} & ResNet-101 & 43.77 & 55.46 & - \\
CCNet~\cite{huang2019ccnet} & ResNet-101 & 45.22 & - & - \\
OCNet~\cite{yuan2018ocnet} & ResNet-101 & 45.45 & - & - \\
SVCNet~\cite{ding2019svcnet} & ResNet-101  & - & - & 53.2  \\
EMANet~\cite{li2019emanet} & ResNet-101  & - & - & 53.1  \\
HRNetV2~\cite{sun2019hrnet} & HRNetV2-W48 & 42.99 & - & 54.0 \\
EncNet~\cite{zhang2018encnet} & ResNet-101 & 44.65 & 55.67 & 52.6 \\
DANet~\cite{jun2019danet} & ResNet-101  & 45.22 & - & 52.6  \\
CFNet~\cite{zhang2019cfnet} & ResNet-101 & 44.89 & - & 54.0 \\
ANN~\cite{zhu2019ann} & ResNet-101 & 45.24 & - & 52.8 \\
DMNet~\cite{he2019dmnet} & ResNet-101 & 45.50 & - & 54.4 \\
ACNet~\cite{fu2019adaptive} & ResNet-101  & 45.90 & 55.84 & 54.1 \\
\hline
NL & ResNet-101 & 44.67 & 55.58 & 50.6 \\
DNL (ours) & ResNet-101 & 45.97 & 56.23 & 54.8 \\
\hline
NL & HRNetV2-W48 & 44.82 & 55.60 & 54.2 \\
DNL (ours) & HRNetV2-W48 & 45.82 & 55.98 & 55.3 \\
\Xhline{1.0pt}
\end{tabular}
}
\label{tab:sota_ade20k}
\vspace{-20pt}
\end{table}

\begin{table}[b]
\small
\centering
    \addtolength{\tabcolsep}{2.5pt}
\caption{Complexity comparisons}
\begin{tabular}{l|ccc}
\Xhline{1.0pt}
& \#param(M)& FLOPs(G) & latency(s/img) \\
\hline
baseline & 70.960 & 691.06 & 0.177 \\
NL&71.484&765.07&0.192\\
DNL&71.485&765.16&0.194\\
\Xhline{1.0pt}
\end{tabular}
\label{tab:complexity}
\end{table}

\noindent \textbf{Complexities.} As discussed in Section \ref{sec.dnl_formulation}, the time and space complexity of the DNL model over the NL model is tiny. Table~\ref{tab:complexity} show the FLOPs and actual latency (single-scale inference using a single GPU) on semantic segmentation, using a ResNet-101 backbone and input resolution of $769\times769$.

\noindent \textbf{Comparison with other methods.}

\noindent \emph{Results on Cityscapes.}
Table~\ref{tab:sota_ccityscapes} shows comparison results for the proposed disentangled non-local network on the Cityscapes test set. Using a ResNet-101 backbone, the disentangled non-local network achieves 82.0\% mIoU, 1.2\% better than that of a standard non-local network. On a stronger backbone of HRNetV2-W48, the disentangled non-local network achieves 0.5\% better accuracy than a standard non-local network. Considering that the standard non-local network has 0.6\% mIoU improvement over a plain HRNetV2-W48 network, such additional gains are significant.

\noindent \emph{Results on ADE20K.}
Table~\ref{tab:sota_ade20k} shows comparison results of the proposed disentangled non-local network on the ADE20k benchmark.
Using a ResNet-101 backbone, the disentangled non-local block achieves 45.97\% and 56.23\% on the validation and test sets, respectively, which are 1.30\% and 0.65\% better than the counterpart networks using a standard non-local block. Our result reveals a new SOTA on this benchmark. On a HRNetV2-W48 backbone, the DNL block is 1.0\% and 0.38\% better than a standard non-local block. Note on ADE20K, HRNetV2-W48 backbone does not perform better than a ResNet-101 backbone, which is different with the other datasets.

\noindent \emph{Results on PASCAL-Context.}
Table~\ref{tab:sota_ccityscapes} shows comparison results of the proposed disentangled non-local network on the PASCAL-Context test set. On ResNet-101, our method improves the standard non-local method significantly, by 3.4\% mIoU (53.7 vs. 50.3). On HRNetV2-W48, our DNL method is 1.1\% mIoU better, which is significant considering that the NL method has 0.2\% improvements over the plain counterpart.

\subsection{Object Detection/Segmentation and Action Recognition}

\textbf{Object Detection and Instance Segmentation on COCO.}
We adopt the open source mmdetection \cite{chen2019mmdetection} codebase for experiments. Following \cite{wang2018non}, the non-local variants are inserted right before the last residual block of c4.

We use the standard configuration of Mask R-CNN \cite{he2017mask} with FPN and ResNet-50 as the backbone architecture, and report the mean average-precision scores at different boxes and the mask IoUs on the COCO2017 validation set.
The input images are resized such that their shorter side is 800 pixels \cite{he2017fpn}.
We trained on 4 GPUs with 4 images per GPU (effective mini batch size of 16).
The backbones of all models are pretrained on ImageNet classification \cite{deng2009imagenet}, then all layers except for stage1 and stage2 are jointly fine-tuned.
In training, synchronized BatchNorm is adopted, and the learning rate scheduler follows the $1\times$ settings of 12 epochs in \cite{he2017mask} where the initial learning rate is 0.02 and decayed by a factor of 10 at the 8$^\text{th}$ and 11$^\text{th}$ epoch. The weight decay is 0.0001 and momentum is 0.9.

Table \ref{table:ablation-coco} shows comparisons of different methods. While the standard non-local block outperforms the baseline counterpart by 0.8\% bbox mAP and 0.7\% mask mAP, the proposed disentangled non-local block brings an additional 0.7\% bbox mAP and 0.6\% mask mAP in gains. Please also see Appendix for experiments when stacking more non-local or disentangled non-local blocks.

\begin{table}[t]
\begin{minipage}[t]{0.6\linewidth}
    \centering
    \footnotesize
    \addtolength{\tabcolsep}{-0.5pt}
\caption{Results based on Mask R-CNN, using R50 as backbone with FPN, for {object detection} and {instance segmentation} on COCO 2017 validation set}
\scalebox{0.9}{
\begin{tabular}{c|ccc|ccc}
\Xhline{1.0pt}
     & AP${^\text{bbox}}$ & AP$^\text{bbox}_\text{50}$ & AP$^\text{bbox}_\text{75}$&AP$^\text{mask}$&AP$^\text{mask}_\text{50}$&AP$^\text{mask}_\text{75}$ \\
\hline
    baseline & 38.8  & 59.3 & 42.5 & 35.1 & 56.2 & 37.9 \\
NL & 39.6 & 60.3 & 43.2 & 35.8 & 57.1 & 38.5 \\
NL$_p$ & 39.8 & 60.4 & 43.7 & 35.9 & 57.3 & 38.4 \\
NL$_u$ & 40.1 & 60.9 & 43.8 & 36.1 & 57.6 & 38.7 \\
DNL & 40.3 & 61.2 & 44.1 & 36.4 & 58.0 & 39.1 \\
\Xhline{1.0pt}
\end{tabular}}
\normalsize
	\label{table:ablation-coco}
\end{minipage}\hfill
\begin{minipage}[t]{0.38\linewidth}
\centering
    \addtolength{\tabcolsep}{2.5pt}
    \footnotesize
\caption{Results based on Slow-only baseline using R50 as backbone on {Kinetics} validation set}
\scalebox{0.9}{
\begin{tabular}{c|cc}
\Xhline{1.0pt}
 & Top-1 Acc & Top-5 Acc \\
\hline
baseline & 74.94 & 91.90 \\
\hline
NL    & 75.95 & 92.29 \\
NL$_p$   & 76.01 & 92.28 \\
NL$_u$   & 75.76 & 92.44 \\
DNL   & 76.31 & 92.69 \\
\Xhline{1.0pt}
\end{tabular}}
\normalsize
\label{table:ablation-kinetics}
\end{minipage}
\vspace{-5pt}
\end{table}

\noindent \textbf{Action Recognition on Kinetics.}
We adopt the Kinetics \cite{kay2017kinetics} dataset for experiments, which includes $\sim$240k training videos and 20k validation videos in 400 human action categories. All models are trained on the training set, and we report the top-1 (\%) and top-5 (\%) accuracy on the validation set.
We adopt the slow-only baseline in \cite{feichtenhofer2018slowfast}, the best single model to date that can utilize weights inflated \cite{carreira2017i3d} from the ImageNet pretrained model. All the experiment settings follow the slow-only baseline in \cite{feichtenhofer2018slowfast}, where 8 frames ($8\times8$) are used as input, and 30-clip validation is adopted. Following~\cite{wang2018non}, we insert (disentangled) non-local blocks after every two residual blocks.

Table \ref{table:ablation-kinetics} shows the comparison of different blocks. It can be seen that the disentangled design performs 0.36\% better than using standard non-local block. Noting the gains of the standard non-local block over the baseline is 1.0, the relative gains of disentangled non-local block over a standard NL block is 36\%.

\section{Conclusion}
In this paper, we first study the non-local block in depth, where we find that its attention computation can be split into two terms, a whitened pairwise term and a unary term. Via both intuitive and statistical analysis, we find that the two terms are tightly coupled in the non-local block, which hinders the learning of each. Based on these findings, we present the disentangled non-local block, where the two terms are decoupled to facilitate learning for both terms. We demonstrate the effectiveness of the decoupled design for learning visual clues on various vision tasks, such as semantic segmentation, object detection and action recognition.

\bibliographystyle{splncs04}
\bibliography{egbib}

\begin{thebibliography}{10}
\providecommand{\url}[1]{\texttt{#1}}
\providecommand{\urlprefix}{URL }
\providecommand{\doi}[1]{https://doi.org/#1}

\bibitem{britz2017massive}
Britz, D., Goldie, A., Luong, M.T., Le, Q.: Massive exploration of neural
  machine translation architectures. arXiv preprint arXiv:1703.03906  (2017)

\bibitem{cao2019gcnet}
Cao, Y., Xu, J., Lin, S., Wei, F., Hu, H.: Gcnet: Non-local networks meet
  squeeze-excitation networks and beyond. arXiv preprint arXiv:1904.11492
  (2019)

\bibitem{carreira2017i3d}
Carreira, J., Zisserman, A.: Quo vadis, action recognition? a new model and the
  kinetics dataset. In: proceedings of the IEEE Conference on Computer Vision
  and Pattern Recognition. pp. 6299--6308 (2017)

\bibitem{chen2019mmdetection}
Chen, K., Wang, J., Pang, J., Cao, Y., Xiong, Y., Li, X., Sun, S., Feng, W.,
  Liu, Z., Xu, J., et~al.: Mmdetection: Open mmlab detection toolbox and
  benchmark. arXiv preprint arXiv:1906.07155  (2019)

\bibitem{chen2017deeplabv3}
Chen, L.C., Papandreou, G., Schroff, F., Adam, H.: Rethinking atrous
  convolution for semantic image segmentation. arXiv preprint arXiv:1706.05587
  (2017)

\bibitem{Chen_2020_CVPR}
Chen, Y., Cao, Y., Hu, H., Wang, L.: Memory enhanced global-local aggregation
  for video object detection. In: The Conference on Computer Vision and Pattern
  Recognition (CVPR) (June 2020)

\bibitem{cheng2019spgnet}
Cheng, B., Chen, L.C., Wei, Y., Zhu, Y., Huang, Z., Xiong, J., Huang, T.S.,
  Hwu, W.M., Shi, H.: Spgnet: Semantic prediction guidance for scene parsing.
  In: Proceedings of the IEEE International Conference on Computer Vision. pp.
  5218--5228 (2019)

\bibitem{cordts2016cityscapes}
Cordts, M., Omran, M., Ramos, S., Rehfeld, T., Enzweiler, M., Benenson, R.,
  Franke, U., Roth, S., Schiele, B.: The cityscapes dataset for semantic urban
  scene understanding. In: Proceedings of the IEEE conference on computer
  vision and pattern recognition. pp. 3213--3223 (2016)

\bibitem{deng2009imagenet}
Deng, J., Dong, W., Socher, R., Li, L.J., Li, K., Fei-Fei, L.: Imagenet: A
  large-scale hierarchical image database. In: 2009 IEEE conference on computer
  vision and pattern recognition. pp. 248--255. Ieee (2009)

\bibitem{Deng_2019_ICCV}
Deng, J., Pan, Y., Yao, T., Zhou, W., Li, H., Mei, T.: Relation distillation
  networks for video object detection. In: The IEEE International Conference on
  Computer Vision (ICCV) (October 2019)

\bibitem{ding2019svcnet}
Ding, H., Jiang, X., Shuai, B., Liu, A.Q., Wang, G.: Semantic correlation
  promoted shape-variant context for segmentation. In: Proceedings of the IEEE
  Conference on Computer Vision and Pattern Recognition. pp. 8885--8894 (2019)

\bibitem{feichtenhofer2018slowfast}
Feichtenhofer, C., Fan, H., Malik, J., He, K.: Slowfast networks for video
  recognition. arXiv preprint arXiv:1812.03982  (2018)

\bibitem{jun2019danet}
Fu, J., Liu, J., Tian, H., Li, Y., Bao, Y., Fang, Z., Lu, H.: Dual attention
  network for scene segmentation. In: Proceedings of the IEEE Conference on
  Computer Vision and Pattern Recognition. pp. 3146--3154 (2019)

\bibitem{fu2019adaptive}
Fu, J., Liu, J., Wang, Y., Li, Y., Bao, Y., Tang, J., Lu, H.: Adaptive context
  network for scene parsing. In: Proceedings of the IEEE international
  conference on computer vision. pp. 6748--6757 (2019)

\bibitem{gu2018learning}
Gu, J., Hu, H., Wang, L., Wei, Y., Dai, J.: Learning region features for object
  detection. In: Proceedings of the European Conference on Computer Vision
  (ECCV). pp. 381--395 (2018)

\bibitem{Guo_2019_ICCV}
Guo, C., Fan, B., Gu, J., Zhang, Q., Xiang, S., Prinet, V., Pan, C.:
  Progressive sparse local attention for video object detection. In: The IEEE
  International Conference on Computer Vision (ICCV) (October 2019)

\bibitem{he2019dmnet}
He, J., Deng, Z., Qiao, Y.: Dynamic multi-scale filters for semantic
  segmentation. In: Proceedings of the IEEE International Conference on
  Computer Vision. pp. 3562--3572 (2019)

\bibitem{he2017mask}
He, K., Gkioxari, G., Doll{\'a}r, P., Girshick, R.: Mask r-cnn. In: Proceedings
  of the IEEE international conference on computer vision. pp. 2961--2969
  (2017)

\bibitem{he2016deep}
He, K., Zhang, X., Ren, S., Sun, J.: Deep residual learning for image
  recognition. In: Proceedings of the IEEE conference on computer vision and
  pattern recognition. pp. 770--778 (2016)

\bibitem{hoshen2017vain}
Hoshen, Y.: Vain: Attentional multi-agent predictive modeling. In: Advances in
  Neural Information Processing Systems. pp. 2701--2711 (2017)

\bibitem{hu2017relation}
Hu, H., Gu, J., Zhang, Z., Dai, J., Wei, Y.: Relation networks for object
  detection (2017)

\bibitem{hu2019local}
Hu, H., Zhang, Z., Xie, Z., Lin, S.: Local relation networks for image
  recognition (2019)

\bibitem{huang2019ccnet}
Huang, Z., Wang, X., Huang, L., Huang, C., Wei, Y., Liu, W.: Ccnet: Criss-cross
  attention for semantic segmentation. In: Proceedings of the IEEE
  International Conference on Computer Vision. pp. 603--612 (2019)

\bibitem{kay2017kinetics}
Kay, W., Carreira, J., Simonyan, K., Zhang, B., Hillier, C., Vijayanarasimhan,
  S., Viola, F., Green, T., Back, T., Natsev, P., et~al.: The kinetics human
  action video dataset. arXiv preprint arXiv:1705.06950  (2017)

\bibitem{li2019emanet}
Li, X., Zhong, Z., Wu, J., Yang, Y., Lin, Z., Liu, H.: Expectation-maximization
  attention networks for semantic segmentation. In: Proceedings of the IEEE
  International Conference on Computer Vision. pp. 9167--9176 (2019)

\bibitem{he2017fpn}
Lin, T.Y., Dollar, P., Girshick, R., He, K., Hariharan, B., Belongie, S.:
  Feature pyramid networks for object detection. In: The IEEE Conference on
  Computer Vision and Pattern Recognition (CVPR) (July 2017)

\bibitem{long2015fcn}
Long, J., Shelhamer, E., Darrell, T.: Fully convolutional networks for semantic
  segmentation. In: Proceedings of the IEEE conference on computer vision and
  pattern recognition. pp. 3431--3440 (2015)

\bibitem{mottaghi2014the}
Mottaghi, R., Chen, X., Liu, X., Cho, N.G., Lee, S.W., Fidler, S., Urtasun, R.,
  Yuille, A.: The role of context for object detection and semantic
  segmentation in the wild. In: IEEE Conference on Computer Vision and Pattern
  Recognition (CVPR) (2014)

\bibitem{pang2019seenet}
Pang, Y., Li, Y., Shen, J., Shao, L.: Towards bridging semantic gap to improve
  semantic segmentation. In: Proceedings of the IEEE International Conference
  on Computer Vision. pp. 4230--4239 (2019)

\bibitem{santoro2017simple}
Santoro, A., Raposo, D., Barrett, D.G., Malinowski, M., Pascanu, R., Battaglia,
  P., Lillicrap, T.: A simple neural network module for relational reasoning.
  In: Advances in neural information processing systems. pp. 4967--4976 (2017)

\bibitem{sun2019hrnet}
Sun, K., Zhao, Y., Jiang, B., Cheng, T., Xiao, B., Liu, D., Mu, Y., Wang, X.,
  Liu, W., Wang, J.: High-resolution representations for labeling pixels and
  regions. arXiv preprint arXiv:1904.04514  (2019)

\bibitem{tang2018analysis}
Tang, G., Sennrich, R., Nivre, J.: An analysis of attention mechanisms: The
  case of word sense disambiguation in neural machine translation. arXiv
  preprint arXiv:1810.07595  (2018)

\bibitem{vaswani2017attention}
Vaswani, A., Shazeer, N., Parmar, N., Uszkoreit, J., Jones, L., Gomez, A.N.,
  Kaiser, {\L}., Polosukhin, I.: Attention is all you need. In: Advances in
  neural information processing systems. pp. 5998--6008 (2017)

\bibitem{wang2018non}
Wang, X., Girshick, R., Gupta, A., He, K.: Non-local neural networks. In:
  Proceedings of the IEEE Conference on Computer Vision and Pattern
  Recognition. pp. 7794--7803 (2018)

\bibitem{watters2017visual}
Watters, N., Zoran, D., Weber, T., Battaglia, P., Pascanu, R., Tacchetti, A.:
  Visual interaction networks: Learning a physics simulator from video. In:
  Advances in neural information processing systems. pp. 4539--4547 (2017)

\bibitem{Wu_2019_ICCV}
Wu, H., Chen, Y., Wang, N., Zhang, Z.: Sequence level semantics aggregation for
  video object detection. In: The IEEE International Conference on Computer
  Vision (ICCV) (October 2019)

\bibitem{Xu_2019_ICCV}
Xu, J., Cao, Y., Zhang, Z., Hu, H.: Spatial-temporal relation networks for
  multi-object tracking. In: The IEEE International Conference on Computer
  Vision (ICCV) (October 2019)

\bibitem{yang2018denseaspp}
Yang, M., Yu, K., Zhang, C., Li, Z., Yang, K.: Denseaspp for semantic
  segmentation in street scenes. In: Proceedings of the IEEE Conference on
  Computer Vision and Pattern Recognition. pp. 3684--3692 (2018)

\bibitem{yuan2018ocnet}
Yuan, Y., Wang, J.: Ocnet: Object context network for scene parsing. arXiv
  preprint arXiv:1809.00916  (2018)

\bibitem{zhang2019acfnet}
Zhang, F., Chen, Y., Li, Z., Hong, Z., Liu, J., Ma, F., Han, J., Ding, E.:
  Acfnet: Attentional class feature network for semantic segmentation. In:
  Proceedings of the IEEE International Conference on Computer Vision. pp.
  6798--6807 (2019)

\bibitem{zhang2018encnet}
Zhang, H., Dana, K., Shi, J., Zhang, Z., Wang, X., Tyagi, A., Agrawal, A.:
  Context encoding for semantic segmentation. In: Proceedings of the IEEE
  Conference on Computer Vision and Pattern Recognition. pp. 7151--7160 (2018)

\bibitem{zhang2019cfnet}
Zhang, H., Zhang, H., Wang, C., Xie, J.: Co-occurrent features in semantic
  segmentation. In: Proceedings of the IEEE Conference on Computer Vision and
  Pattern Recognition. pp. 548--557 (2019)

\bibitem{zhao2017pspnet}
Zhao, H., Shi, J., Qi, X., Wang, X., Jia, J.: Pyramid scene parsing network.
  In: Proceedings of the IEEE conference on computer vision and pattern
  recognition. pp. 2881--2890 (2017)

\bibitem{zhao2018psanet}
Zhao, H., Zhang, Y., Liu, S., Shi, J., Change~Loy, C., Lin, D., Jia, J.:
  Psanet: Point-wise spatial attention network for scene parsing. In:
  Proceedings of the European Conference on Computer Vision (ECCV). pp.
  267--283 (2018)

\bibitem{zhou2017scene}
Zhou, B., Zhao, H., Puig, X., Fidler, S., Barriuso, A., Torralba, A.: Scene
  parsing through ade20k dataset. In: Proceedings of the IEEE conference on
  computer vision and pattern recognition. pp. 633--641 (2017)

\bibitem{zhu2019empirical}
Zhu, X., Cheng, D., Zhang, Z., Lin, S., Dai, J.: An empirical study of spatial
  attention mechanisms in deep networks. In: The IEEE International Conference
  on Computer Vision (ICCV) (October 2019)

\bibitem{zhu2019ann}
Zhu, Z., Xu, M., Bai, S., Huang, T., Bai, X.: Asymmetric non-local neural
  networks for semantic segmentation. In: Proceedings of the IEEE International
  Conference on Computer Vision. pp. 593--602 (2019)

\end{thebibliography}

\clearpage
\appendix

\begin{table}
    \centering
    \footnotesize
    \addtolength{\tabcolsep}{-0.5pt}
\caption{Results with more NL and DNL blocks based on Mask R-CNN, using R50 as backbone with FPN, for {object detection} and {instance segmentation} on COCO 2017 validation set}
\scalebox{0.9}{
\begin{tabular}{c|ccc|ccc}
\Xhline{1.0pt}
     & AP${^\text{bbox}}$ & AP$^\text{bbox}_\text{50}$ & AP$^\text{bbox}_\text{75}$&AP$^\text{mask}$&AP$^\text{mask}_\text{50}$&AP$^\text{mask}_\text{75}$ \\
\hline
    baseline & 38.8  & 59.3 & 42.5 & 35.1 & 56.2 & 37.9 \\
    \hline
NL (c4 one) & 39.6 & 60.3 & 43.2 & 35.8 & 57.1 & 38.5 \\
NL (c5 all) & 40.0 & 62.1 & 43.5 & 36.1 & 58.6 & 38.6 \\
NL (c4c5 all) & 40.1 & 62.3 & 43.5 & 36.0 & 58.9 & 38.3 \\
\hline
DNL (c4 one) & 40.3 & 61.2 & 44.1 & 36.4 & 58.0 & 39.1 \\
DNL (c5 all) & 41.2 & 62.7 & 44.7 & 37.0 & 59.5 & 39.5 \\
DNL (c4c5 all) & 41.4 & 63.2 & 45.3 & 37.3 & 59.8 & 39.8 \\
\Xhline{1.0pt}
\end{tabular}}
\normalsize
\label{tab:ablation-coco}
\end{table}

\section{More NL/DNL blocks for COCO Object Detection}

In section 5.2 of the main paper, we follow the settings in~\cite{wang2018non} where 1 non-local (NL) or disentangled non-local (DNL) block is inserted right before the last residual block of c4. In this section, we investigate the performance of NL and DNL when more attention blocks are inserted into the backbone, as shown in Table~\ref{tab:ablation-coco}.

While the proposed DNL method outperforms NL method by 0.7\% bbox mAP and 0.6\% mask mAP when 1 attention block is inserted into the backbone (denoted as ``c4 one''), the gains brought by the proposed DNL method over the NL method are enlarged to 1.2\% bbox mAP and 0.9\% mask mAP, respectively, when every residual block of stage c5 is followed by 1 attention block (denoted as ``c4 all''). The gains are further enlarged to 1.3\% bbox mAP and 1.3\% mask mAP when additionally every residual block of stage c4 is followed by 1 attention block (denoted as ``c4 c5 all''). These results indicate that the DNL method can benefit more from increasing block number than the NL method.

\section{Detailed Proof of Proposition 1}

The object function $O(\alpha, \beta)$ in Eq.~(3) of the main paper can be rewritten as
\begin{small}
\begin{equation}
\begin{aligned}
O(\alpha, \beta) = \frac{\sum_{i \in \Omega} (\mathbf{q}_i-\mathbf{\alpha})^T A (\mathbf{q}_i-\mathbf{\alpha})}{\sum_{i \in \Omega} \left((\mathbf{q}_i-\mathbf{\alpha})^T (\mathbf{q}_i-\mathbf{\alpha})\right)} + \frac{\sum_{m \in \Omega} (\mathbf{k}_m-\mathbf{\beta})^T B (\mathbf{k}_m-\mathbf{\beta})}{\sum_{m \in \Omega} \left((\mathbf{k}_m-\mathbf{\beta})^T (\mathbf{k}_m-\mathbf{\beta})\right) }
\end{aligned}
\end{equation}
\end{small}where
\begin{small}
\begin{equation}
\begin{aligned}
&A=\frac{\sum_{m,n \in \Omega} (\mathbf{k}_m-\mathbf{k}_n)(\mathbf{k}_m-\mathbf{k}_n)^T}{\sum_{m,n \in \Omega} (\mathbf{k}_m-\mathbf{k}_n)^T (\mathbf{k}_m-\mathbf{k}_n)} \qquad
B=\frac{\sum_{i,j \in \Omega} (\mathbf{q}_i-\mathbf{q}_j)(\mathbf{q}_i-\mathbf{q}_j)^T}{\sum_{i,j \in \Omega} (\mathbf{q}_i-\mathbf{q}_j)^T (\mathbf{q}_i-\mathbf{q}_j)}
\end{aligned}
\end{equation}
\end{small}

We first prove that all eigenvalues of matrix $A$ and $B$ are smaller or equal than 1. Denote the eigenvalues of matrix $A$ as $\lambda_{1},..., \lambda_{d}$.
According to Cauchy–Schwarz inequality, we have
\begin{small}
\begin{equation}
\label{eq.lambda}
\begin{aligned}
&\sum_{1 \leqslant i \leqslant d}\lambda_{i}^2=tr(A^T A) \\
&=\text{tr}\left(\frac{\sum_{m,n \in \Omega} (\mathbf{k}_m-\mathbf{k}_n)(\mathbf{k}_m-\mathbf{k}_n)^T  \cdot \sum_{s,t \in \Omega} (\mathbf{k}_s-\mathbf{k}_t)(\mathbf{k}_s-\mathbf{k}_t)^T }{\sum_{m,n \in \Omega} (\mathbf{k}_m-\mathbf{k}_n)^T (\mathbf{k}_m-\mathbf{k}_n) \cdot \sum_{s,t \in \Omega} (\mathbf{k}_s-\mathbf{k}_t^T)(\mathbf{k}_s-\mathbf{k}_t)} \right) \\
&=\frac{\sum_{m,n,s,t \in \Omega} (\mathbf{k}_m-\mathbf{k}_n)^T(\mathbf{k}_s-\mathbf{k}_t)\cdot tr\left((\mathbf{k}_m-\mathbf{k}_n)(\mathbf{k}_s-\mathbf{k}_t)^T\right)}{\left(\sum_{m,n \in \Omega}(\mathbf{k}_m-\mathbf{k}_n)^T (\mathbf{k}_m-\mathbf{k}_n)\right)^2} \\
&=\frac{\sum_{m,n,s,t \in \Omega} \left((\mathbf{k}_m-\mathbf{k}_n)^T(\mathbf{k}_s-\mathbf{k}_t)\right)^2}{\left(\sum_{m,n \in \Omega}(\mathbf{k}_m-\mathbf{k}_n)^T (\mathbf{k}_m-\mathbf{k}_n)\right)^2} \quad \leq 1
\end{aligned}
\end{equation}
\end{small}
Given Eq.~(\ref{eq.lambda}), we have: $\forall 1 \leq i \leq d$, $\lambda_{i} \leq 1$. Similarly, we can prove all eigenvalues of matrix B are smaller or equal than 1. The hessian matrix of Eq.~(1) with respect to $\alpha$ and $\beta$ are non-positive definite matrix. The optimal$\alpha^*$ and $\beta^*$ are thus the solutions of the following equations: $\frac{\partial O} { \partial \alpha} = 0$, $\frac{\partial O} {\partial \beta} = 0$.

For $\alpha^*$, we have
\begin{small}
\begin{equation}
\begin{aligned}
&\frac{\partial O} { \partial \alpha^*} = \sum_{i=1}^{N_{p}}2\left(\frac{\sum_{m,n}(\mathbf{k}_m-\mathbf{k}_n)(\mathbf{k}_m-\mathbf{k}_n)^T}{\sum_{m,n}(\mathbf{k}_m-\mathbf{k}_n)^T(\mathbf{k}_m-\mathbf{k}_n)}-1\right)(\mathbf{q}_i-\alpha^*)=0, \\
& \Leftrightarrow \left(\frac{\sum_{m,n}(\mathbf{k}_m-\mathbf{k}_n)(\mathbf{k}_m-\mathbf{k}_n)^T}{\sum_{m,n}(\mathbf{k}_m-\mathbf{k}_n)^T(\mathbf{k}_m-\mathbf{k}_n)}-1\right)\sum_{i=1}^{N_{p}}2(\mathbf{q}_i-\alpha^*)=0.
\label{con:derive}
\end{aligned}
\end{equation}
\end{small}
To satisfy Eqn.~\ref{con:derive}, we have:
\begin{small}
\begin{equation}
\begin{aligned}
&\sum_{i=1}^{N_{p}}(\mathbf{q}_i-\alpha^*)=0.
\end{aligned}
\end{equation}
\end{small}
The optimal $\alpha^*$ is thus
\begin{small}
\begin{equation}
\begin{aligned}
&\alpha^*=\frac{1}{N_{p}}\sum_{i=1}^{N_{p}}\mathbf{q}_i.
\end{aligned}
\end{equation}
\end{small}Similarly, the optimal $\beta^*$ is computed as
\begin{small}
\begin{equation}
\begin{aligned}
&\beta^*=\frac{1}{N_{p}}\sum_{i=1}^{N_{p}}\mathbf{k}_i.
\end{aligned}
\end{equation}
\end{small}

\section{Proof for Eqn.~4 in the main paper}
Here, we provide a proof for Eqn.~4 in Section 3.1. The dot product of query $\mathbf{q}_i$ and key $\mathbf{k}_j$ can be split into several terms by introducing a whitening operation on the key and query:
\begin{small}
\begin{align}
\mathbf{q}_i^T\mathbf{k}_j=\left(\mathbf{q}_i-\boldsymbol{\mu}_q\right)^T\left(\mathbf{k}_j-\boldsymbol{\mu}_k\right)+\boldsymbol{\mu}_q^T\mathbf{k}_j+\mathbf{q}_i^T\boldsymbol{\mu}_k+\boldsymbol{\mu}_q^T\boldsymbol{\mu}_k,
\end{align}
\end{small}where $\boldsymbol{\mu}_q$ and $\boldsymbol{\mu}_k$ denote $\frac{1}{N_p}\sum_{i=1}^{N_p}\mathbf{q}_i$ and $\frac{1}{N_p}\sum_{i=1}^{N_p}\mathbf{k}_j$, respectively.

Note that the last two terms ($\mathbf{q}_i^T\boldsymbol{\mu}_k$ and $\boldsymbol{\mu}_q^T\boldsymbol{\mu}_k$) are factors in common with both the numerator and denominator of the correlation function $f$ and the normalization factor $\mathcal C$, so they can be eliminated as follows:

\begin{small}
\begin{align}
&\frac{\exp\left(\mathbf{q}_i^T\mathbf{k}_j\right)}{\sum_{t=1}^{N_p}\exp\left(\mathbf{q}_i^T\mathbf{k}_t\right)} \notag\\
&=\frac{\exp\left(\left(\mathbf{q}_i-\boldsymbol{\mu}_q\right)^T\left(\mathbf{k}_j-\boldsymbol{\mu}_k\right)+\boldsymbol{\mu}_q^T\mathbf{k}_j+\mathbf{q}_i^T\boldsymbol{\mu}_k+\boldsymbol{\mu}_q^T\boldsymbol{\mu}_k\right)}{\sum_{t=1}^{N_p}\exp\left(\left(\mathbf{q}_i-\boldsymbol{\mu}_q\right)^T\left(\mathbf{k}_t-\boldsymbol{\mu}_k\right)+\boldsymbol{\mu}_q^T\mathbf{k}_t+\mathbf{q}_i^T\boldsymbol{\mu}_k+\boldsymbol{\mu}_q^T\boldsymbol{\mu}_k\right)} \notag\\
&=\!\frac{\exp\left(\left(\mathbf{q}_i\!-\!\boldsymbol{\mu}_q\right)^T\left(\mathbf{k}_j\!-\!\boldsymbol{\mu}_k\right)\!+\!\boldsymbol{\mu}_q^T\mathbf{k}_j\right)\exp\left(\mathbf{q}_i^T\boldsymbol{\mu}_k\!+\!\boldsymbol{\mu}_q^T\boldsymbol{\mu}_k\right)}{\sum_{t=1}^{N_p}\exp\left(\left(\mathbf{q}_i\!-\!\boldsymbol{\mu}_q\right)^T\left(\mathbf{k}_t\!-\!\boldsymbol{\mu}_k\right)\!+\!\boldsymbol{\mu}_q^T\mathbf{k}_t\right)\exp\left(\mathbf{q}_i^T\boldsymbol{\mu}_k\!+\!\boldsymbol{\mu}_q^T\boldsymbol{\mu}_k\right)} \notag\\
&=\frac{\exp\left(\left(\mathbf{q}_i-\boldsymbol{\mu}_q\right)^T\left(\mathbf{k}_j-\boldsymbol{\mu}_k\right)+\boldsymbol{\mu}_q^T\mathbf{k}_j\right)}{\sum_{t=1}^{N_p}\exp\left(\left(\mathbf{q}_i-\boldsymbol{\mu}_q\right)^T\left(\mathbf{k}_t-\boldsymbol{\mu}_k\right)+\boldsymbol{\mu}_q^T\mathbf{k}_t\right)}.
\end{align}
\end{small}

Finally, we obtain
\begin{small}
\begin{align}
\sigma(\mathbf{q}_i^T\mathbf{k}_j)=\sigma(\underbrace{\left(\mathbf{q}_i-\boldsymbol{\mu}_q\right)^T\left(\mathbf{k}_j-\boldsymbol{\mu}_k\right)}_{pairwise}+\underbrace{\boldsymbol{\mu}_q^T\mathbf{k}_j}_{unary}).
\label{con:dnl_app}
\end{align}
\end{small}

\section{More Examples of Learnt Attention Maps by NL/DNL Methods}

In this section, we show more examples of the learnt attention maps by the NL/DNL methods on the Cityscapes semantic segmentation, COCO object detection/instance segmentation and Kinetics action recognition tasks.

\begin{figure*}[]
\centering
\includegraphics[width=0.95\linewidth]{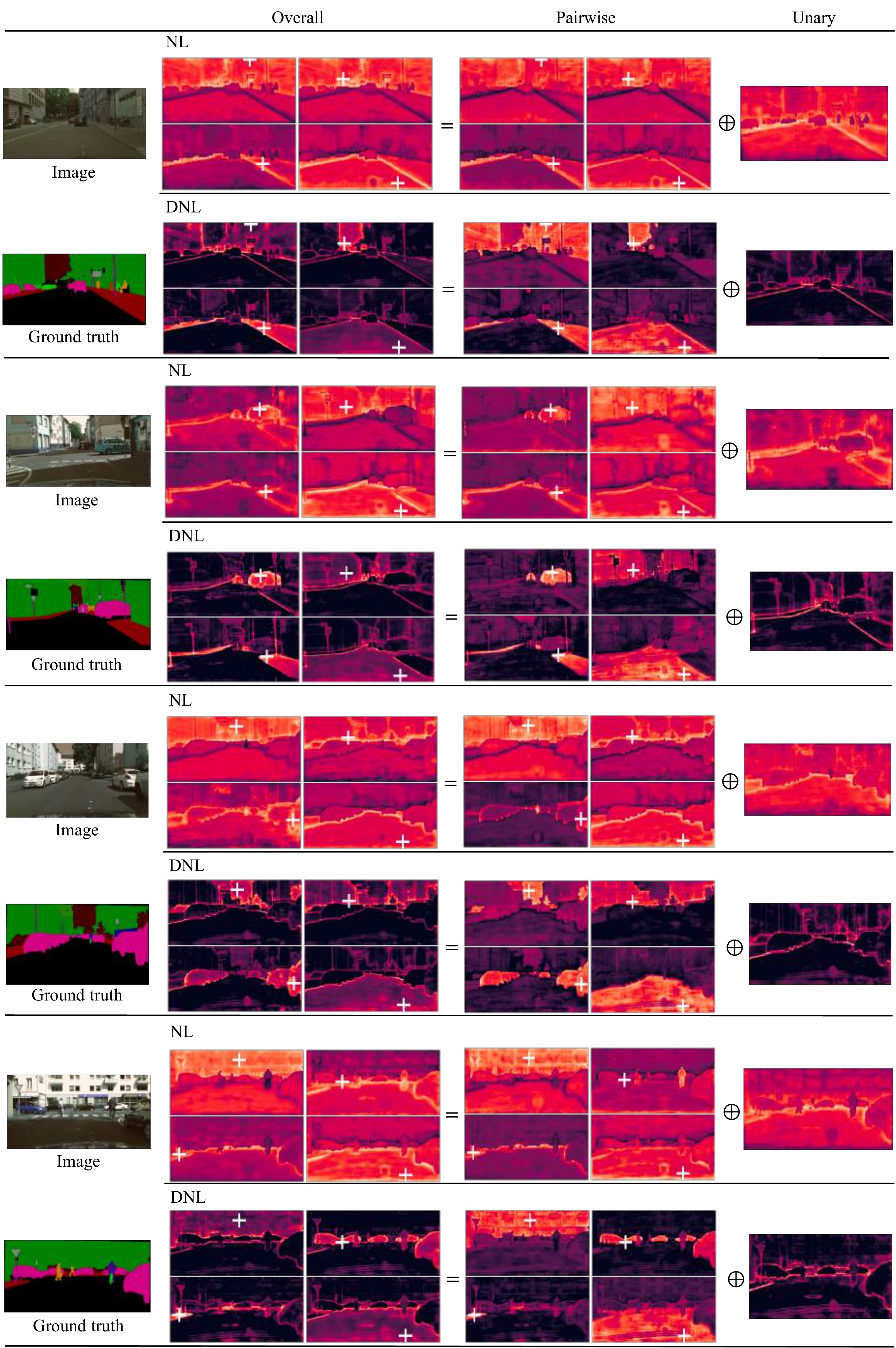}
\caption{Visualization of attention maps of NL block and our DNL block on Cityscapes Dataset. The query points are marked in white cross}
\label{fig:city_appendix}
\end{figure*}

Fig. \ref{fig:city_appendix} show more examples of the learnt attention maps by NL/DNL on Cityscapes. With DNL block, the whitened pairwise term learns clear within-region clues while the unary term learns salient boundaries, which cannot be observed in that of the original NL block.

\begin{figure*}[]
\centering
\includegraphics[width=0.95\linewidth]{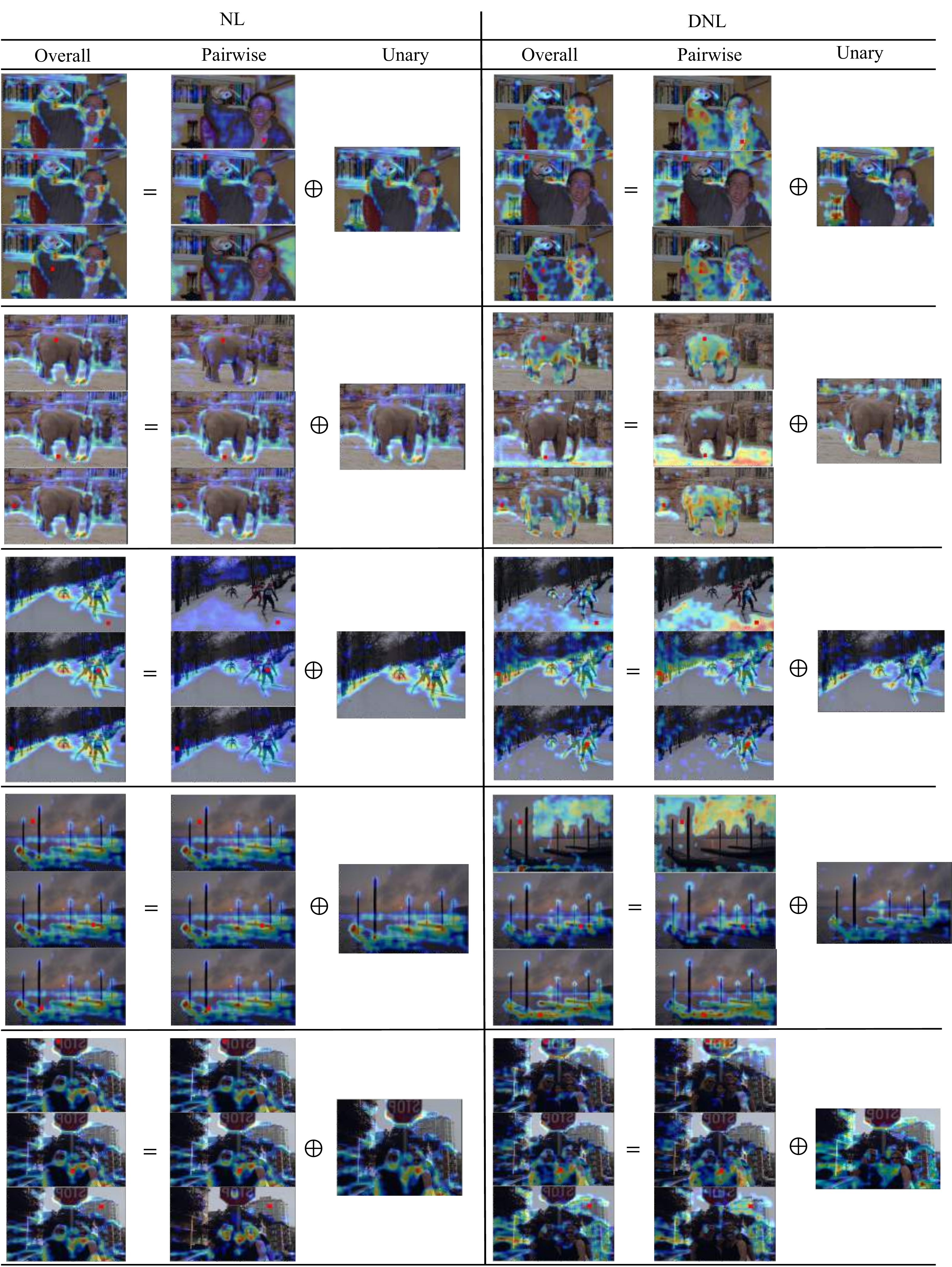}
\caption{Visualization of attention maps of NL block and our DNL block on COCO object detection task. The query points are marked in red.}
\label{fig:det_appendix}
\end{figure*}

Fig. \ref{fig:det_appendix} show more examples of the learnt attention maps by NL/DNL on COCO object detection/instance segmentation.
It can be seen that the attention maps of NL block are mainly dominated by the unary term that different query points (marked in red) have similar overall attention maps. In DNL, the pariwise term in DNL shows clear within-region meaning and appears significant in the final overall attention maps, that different query points have different overall attention maps.
DNL also shows more focus to salient regions  than the one in an NL block.

\begin{figure*}[]
\centering
\includegraphics[width=1.0\linewidth]{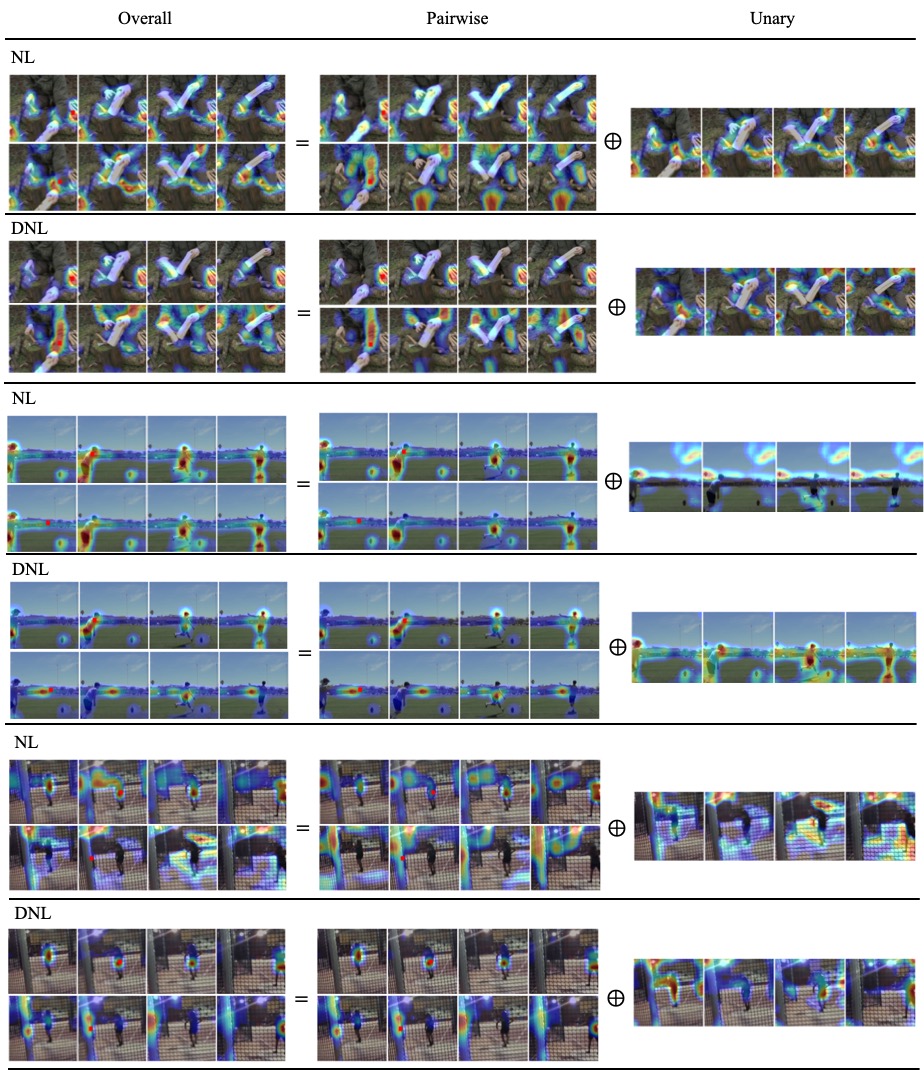}
\caption{Visualization of attention maps of NL block and our DNL block on Kinetics action recognition. 4 frames of a video clip are visualized. For each sample of each block, two different queries are chosen as the top and bottom rows. The query points are marked in red}
\label{fig:att_action}
\end{figure*}

Fig. \ref{fig:att_action} show more examples of the learnt attention maps by NL/DNL on Kinetics action recognition task. 4 frames in a video clip are visualized. The unary term of DNL shows better focus to salient regions than the one in an NL block. The pairwise term in DNL also shows clearer within-region meaning than that in an NL block.


\end{document}